\documentclass[english,british]{article}
\usepackage[T1]{fontenc}
\usepackage[latin9]{inputenc}
\usepackage{array}
\usepackage{float}
\usepackage{multirow}
\usepackage{amsmath}
\usepackage{amssymb}
\usepackage{stmaryrd}
\usepackage{graphicx}
\usepackage{setspace}

\makeatletter

\providecommand{\tabularnewline}{\\}
\floatstyle{ruled}
\newfloat{algorithm}{tbp}{loa}
\providecommand{\algorithmname}{Algorithm}
\floatname{algorithm}{\protect\algorithmname}

\usepackage{aaai19}
\usepackage{algorithmic}

\@ifundefined{showcaptionsetup}{}{%
 \PassOptionsToPackage{caption=false}{subfig}}
\usepackage{subfig}
\makeatother

\usepackage{babel}
\addto\captionsenglish{\renewcommand{\algorithmname}{Algorithm}}

\begin{document}
\title{Bayesian functional optimisation with shape prior} 
\author{\\Pratibha Vellanki, Santu Rana, Sunil Gupta, David Rubin de Celis Leal\\
Alessandra Sutti, Murray Height, Svetha Venkatesh\\
Centre for Pattern Recognition and Data Analytics\\
Deakin University, Geelong, Australia\\
pratibha.vellanki, santu.rana, sunil.gupta, svetha.venkatesh @deakin.edu.au\\
Institute for Frontier Materials, GTP Research\\
Deakin University, Geelong, Australia\\
d.rubindecelisleal, alessandra.sutti, murray.height@deakin.edu.au} 
\maketitle

\selectlanguage{english}%

\global\long\def\mF{\mathcal{F}}

\global\long\def\mA{\mathcal{A}}

\global\long\def\mH{\mathcal{H}}

\global\long\def\mX{\mathcal{X}}

\global\long\def\dist{d}

\global\long\def\HX{\entro\left(X\right)}
 \global\long\def\entropyX{\HX}

\global\long\def\HY{\entro\left(Y\right)}
 \global\long\def\entropyY{\HY}

\global\long\def\HXY{\entro\left(X,Y\right)}
 \global\long\def\entropyXY{\HXY}

\global\long\def\mutualXY{\mutual\left(X;Y\right)}
 \global\long\def\mutinfoXY{W\mutualXY}

\global\long\def\given{\mid}

\global\long\def\gv{\given}

\global\long\def\goto{\rightarrow}

\global\long\def\asgoto{\stackrel{a.s.}{\longrightarrow}}

\global\long\def\pgoto{\stackrel{p}{\longrightarrow}}

\global\long\def\dgoto{\stackrel{d}{\longrightarrow}}

\global\long\def\ll{\mathit{l}}

\global\long\def\logll{\mathcal{L}}

\global\long\def\bzero{\vt0}

\global\long\def\bone{\mathbf{1}}

\global\long\def\bff{\vt f}

\global\long\def\bx{\boldsymbol{x}}

\global\long\def\bX{\boldsymbol{X}}

\global\long\def\bW{\mathbf{W}}

\global\long\def\bH{\mathbf{H}}

\global\long\def\bL{\mathbf{L}}

\global\long\def\tbx{\tilde{\bx}}

\global\long\def\by{\boldsymbol{y}}

\global\long\def\bY{\boldsymbol{Y}}

\global\long\def\bz{\boldsymbol{z}}

\global\long\def\bZ{\boldsymbol{Z}}

\global\long\def\bu{\boldsymbol{u}}

\global\long\def\bU{\boldsymbol{U}}

\global\long\def\bv{\boldsymbol{v}}

\global\long\def\bV{\boldsymbol{V}}

\global\long\def\bw{\vt w}

\global\long\def\balpha{\gvt\alpha}

\global\long\def\bbeta{\gvt\beta}

\global\long\def\bmu{\gvt\mu}

\global\long\def\btheta{\boldsymbol{\theta}}

\global\long\def\blambda{\boldsymbol{\lambda}}

\global\long\def\realset{\mathbb{R}}

\global\long\def\realn{\real^{n}}

\global\long\def\natset{\integerset}

\global\long\def\interger{\integerset}

\global\long\def\integerset{\mathbb{Z}}

\global\long\def\natn{\natset^{n}}

\global\long\def\rational{\mathbb{Q}}

\global\long\def\realPlusn{\mathbb{R_{+}^{n}}}

\global\long\def\comp{\complexset}
 \global\long\def\complexset{\mathbb{C}}

\global\long\def\and{\cap}

\global\long\def\compn{\comp^{n}}

\global\long\def\comb#1#2{\left({#1\atop #2}\right) }
\selectlanguage{british}%

\title{Bayesian functional optimisation with shape prior}
\maketitle
\begin{abstract}
Real world experiments are expensive, and thus it is important to
reach a target in minimum number of experiments. Experimental processes
often involve control variables that changes over time. Such problems
can be formulated as a functional optimisation problem. We develop
a novel Bayesian optimisation framework for such functional optimisation
of expensive black-box processes. We represent the control function
using Bernstein polynomial basis and optimise in the coefficient space.
We derive the theory and practice required to dynamically adjust the
order of the polynomial degree, and show how prior information about
shape can be integrated. We demonstrate the effectiveness of our approach
for short polymer fibre design and optimising learning rate schedules
for deep networks.
\end{abstract}

\section{Introduction}

Functional optimisation arises when a time-varying system requires
optimal control variable values to change with time. As an example
consider optimising the learning rate schedule while training a neural
network. The learning rate schedule can be expressed as a function
of time and often there is a about the shape of this function; traditionally,
it is a decreasing function. Consider also recirculation processes,
common in industries like drug production or plastic recycling. Recirculation
involves the reintroduction of partially formed output product into
the input of the system until target output is reached. Recirculation
might require adjusting of input parameters as a function of time
to keep the system optimal throughout. Often, industrial knowledge
exists about the trend of this adjustment.

We propose a Bayesian functional optimisation algorithm for expensive
processes that offers two main capabilities: a) allows detection of
underspecification of complexity of the functional search space and
adjusting for it in a dynamic fashion, and b) admits loose prior information
on the shape of the function.

The closest work is that of \cite{vien2018bayesian}, where functions
are represented in a functional RKHS, captured through functional
kernels. Based on that a Gaussian process is constructed and subsequently,
Bayesian optimisation is performed. However, their method does not
allow incorporating prior information or for adjustment of function
complexity mid-optimisation.

Our solution is primarily based on representing the control function
on Bernstein polynomial basis \cite{bernstein1912demonstration} and
then optimising on the coefficient space. Bernstein polynomials basis
follows the Stone-Weierstrass approximation theorem i.e. any function
on a bounded subspace when represented on this basis system can be
point-wise approximated to an arbitrary precision. Whilst the theorem
is true for any polynomial basis system, Bernstein basis offers many
unique properties, some of which are critical for achieving our goals.
Once the control functions are represented on a Bernstein basis with
a suitable order we can directly use the coefficient vector as the
input subspace for the global optimisation of a function which maps
the coefficient vector to the outcome of the system.

Bernstein polynomials have been a popular choice in the field of aerospace
for the optimisation of aerofoil geometry. For example, it has been
used in \cite{kulfan2006fundamental} to convert a shape optimisation
problem into a function optimisation problem. Often Computational
Fluid Dynamic (CFD) tools are used in to optimise such geometries
\cite{samareh2001survey}. However, these methods are not designed
to be sample-efficient, and hence, not feasible for expensive optimisation
tasks. The proposed Bayesian functional optimisation algorithm addresses
the optimisation of expensive experimental processes, two of which
we present in this paper (physical recirculation systems, and optimisation
of learning rate schedule for a large neural network model being trained
on a large dataset). 

Mathematically, if {\small{}$g(t)$} is the control function that
drives the system output ($y$) by a functional {\small{}$h:g(t)\rightarrow y$}
where {\small{}$g\in B(R)$}, the space of all bounded real-valued
functions, then the original functional optimisation problem can be
written as: {\small{}
\[
g^{*}(t)=\text{argmax}_{g(t)\in B(R)}h(g(t))
\]
}When we convert {\small{}$g(t)$} onto the $n$th order Bernstein
polynomial as\emph{ }\emph{\small{}$g(t)=\sum_{v=0}^{n}\alpha_{v}b_{v,n}(t)$}\emph{,
}where $\boldsymbol{\alpha}=\{\alpha_{v}\}_{v=1}^{n}$are the Bernstein
coefficients and{\small{} $b_{v,n}(t)$ }are the base polynomials
(more in framework), the above optimisation problem can thus be reformulated
in a more familiar function optimisation problem as:{\small{}
\begin{align*}
\boldsymbol{\alpha}^{*} & =\text{argmax}_{\alpha\in\mathcal{A}}f(\boldsymbol{\alpha})
\end{align*}
}where {\small{}$f:\boldsymbol{\alpha}\rightarrow y$}. In addition
to a control function, we may also have some other control variables
({\small{}$\boldsymbol{u}$}) that need to be optimised. For example,
in the case of neural network hyper-parameter tuning, {\small{}$\boldsymbol{u}$}
may represent network size parameters, whilst {\small{}$\boldsymbol{\alpha}$}
represents the learning rate schedule. In such cases, our formulation
can be easily extended to{\small{} $\{\boldsymbol{\alpha}^{*},\boldsymbol{u^{*}}\}=\text{argmax}_{\alpha\in\mathcal{A},u\in\mathcal{U}}f(\boldsymbol{\alpha},\boldsymbol{u})$.
}Putting them together, we will henceforth indicate{\small{} $\boldsymbol{x}=\{\boldsymbol{\alpha},\boldsymbol{u}\}$,
}and{\small{} $\mathcal{X}=\mathcal{A}\cup\mathcal{U}$.}{\small\par}

\begin{figure}
\begin{centering}
\includegraphics[width=1\columnwidth]{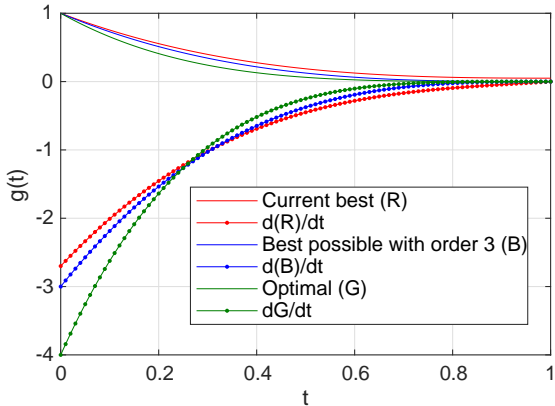}
\par\end{centering}
\caption{{\small{}Intuition behind determining the whether the current order
of Bernstein suffices to capture the optimal function.\label{fig:derivative}}}
\end{figure}

Next, following the properties of derivative of function on Bernstein
basis, and using results from \cite{chang2007shape}, we show that
prior information on shapes such as monotonicity or unimodality can
be encoded by simply adding constraint functions on the values of
{\small{}$\boldsymbol{\alpha}$}. We further propose a principled
way to detect if the order of the polynomial is underspecified. The
intuition is as follows: Figure \ref{fig:derivative} shows three
functions at the top and their corresponding derivative at the bottom
panel with matching colour. Let us assume that the current function
we have with a 3rd order Bernstein polynomial is the red function.
The maximum derivative of the red function happens at $t=0$ and it
is around -2.8. The target function is depicted via the colour green
and its maximum derivative as seen from the bottom panel is -4 at
$t=0$. In the absolute sense the derivative magnitude is higher than
the current function that we have. However, we see that the derivative
limit of a 3rd order Bernstein polynomial (Blue function) is at -3,
which in the magnitude sense lower than the maximum derivative of
the target function (green). Thus the optimal function (green) cannot
be reached by a 3rd order Bernstein polynomial. The target function
can only be reached if the order is increased. An underspecification
of the order is thus detected if the derivative magnitude is close
to the theoretical maxima possible within the current order of the
polynomial basis based on the ranges over \foreignlanguage{english}{$\boldsymbol{\alpha}$}.
When an underspecification is detected, we increment the order of
the Bernstein basis. We derive a key theorem to compute the maximum
of the derivative for a function realised using Bernstein polynomial
basis of fixed order. Next, using the existing results of order elevation
for Bernstein basis, we show how to reuse observation obtained using
lower order basis for the new higher order basis. In some cases, it
may not be possible to detect order underspecification via derivative
checking (if the function has more number of modes than can be modelled
by the current order of the polynomial), and hence, we also increment
order at a fixed interval up to a maximum specified order. In all
time, we use all the observations using order elevation technique.
Convergence is guaranteed as long as the function is realisable within
the maximum order specified. 

\begin{figure}
\begin{centering}
\includegraphics[width=1\columnwidth]{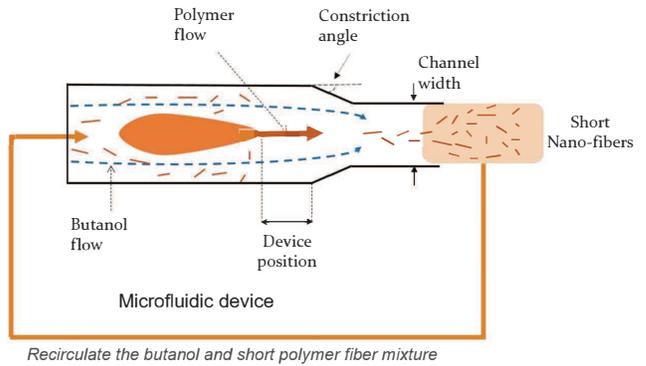}
\par\end{centering}
\caption{{\small{}Short polymer fibre production using recirculation.\label{fig:Recirculation}}}
\end{figure}

We apply our algorithm to two problems: the design and production
of concentrated short polymer fibre solution using recirculation and
learning rate schedule optimisation for neural network training. Interesting
new materials like short polymer fibres can impart exotic properties
to natural fabrics \cite{feng2003creation,ma2008review}. Production
involves injecting a liquid polymer into a high speed butanol flow
through a specially designed apparatus (Figure \ref{fig:Recirculation}).
This turns the liquid polymer into short nano-scale fibres. The control
variables include the apparatus geometry and flow rates which in turn
determine the produced fibre quality. To increase the fibre concentration
in the mixture produced, the same mixture (butanol+fibre) is recirculated
through the apparatus, keeping the polymer flow uninterrupted. Since
the recirculation process introduces dynamics in the constituents
of the mixture, one may need to change the control variables to keep
them optimal throughout this dynamic process. We apply our algorithm
in maximising the quality of fibre yield for the already mentioned
short polymer fibre production process. During recirculation we only
change the butanol flow rate as a function of time, as others are
not easy to change dynamically in the used setup. We used the experimenter's
hunch that an increasing flow rate will result in the highest quality
fibre. In our experiments we found two profiles, one which is nearly
constant, and the other which is increasing that both result in the
highest quality possible, loosely validating the experimenter's hunch. 

In neural network training, the learning rate schedule can be modelled
either as a long vector or as a function of epochs \cite{bengio2012practical}.
The latter is attractive as the smoothness in the consecutive learning
rate values implies smaller effective design space when considered
as a function than with a full blown vector of the corresponding schedule.
We  apply our algorithm for learning rate schedule optimisation and
found that an optimised learning rate schedule can even make SGD to
perform better than both the method proposed by \cite{vien2018bayesian}
and a state of the art optimiser with automatic scheduling like Adam.

\section{Bayesian optimisation}

Bayesian optimisation is a global optimisation method for expensive
black-box function \cite{jones1998efficient,mockus1994application}.
The optimisation problem:
\begin{center}
$\boldsymbol{x}^{*}=\text{argmax}_{\boldsymbol{x}\in\mathcal{X}}f(\boldsymbol{x})$
\par\end{center}

\noindent The function is usually modelled using a \emph{Gaussian
Process }\cite{rasmussen2006gaussian} as the prior \emph{i.e.}{\small{}
\begin{eqnarray*}
f(\boldsymbol{x}) & \sim & \mathcal{GP}(m(\boldsymbol{x}),k(\boldsymbol{x},\boldsymbol{x}')).
\end{eqnarray*}
}where{\small{} $m(\boldsymbol{x})$} and {\small{}$k(\boldsymbol{x},\boldsymbol{x}')$
}are the mean and the covariance function of the Gaussian process
\cite{Brochu_Cora_Freitas_2010Tutorial}. Mean function {\small{}$m(\boldsymbol{x})$}
can be assumed to be a zero function without any loss of generalisation.
Popular covariance functions include squared exponential (SE) kernel,
Mat\'ern kernel etc. The predictive mean and variance of the Gaussian
process is a Gaussian distribution, whose encapsulates epistemic uncertainty.
Using an observation model of{\small{} $y=f(\boldsymbol{x})+\epsilon$},
where {\small{}$\epsilon\sim\mathcal{N}(0,\sigma_{noise}^{2})$},
and denoting {\small{}$\mathcal{D}=\{\boldsymbol{x}_{i},y_{i}\}_{i=1}^{t}$}
one can derive the predictive distribution as:
\begin{center}
$P(f_{t+1}\given\mathcal{D}_{1:t},\boldsymbol{x})=\mathcal{N}(\mu_{t+1}(\boldsymbol{x}),\sigma_{t+1}^{2}(\boldsymbol{x}))$
\par\end{center}

\noindent where,{\small{} $\boldsymbol{k}=[k(\boldsymbol{x},\boldsymbol{x}_{1}),\ldots,k(\boldsymbol{x},\boldsymbol{x}_{t})]$},
the kernel matrix{\small{} $[K_{ij}]=k(\boldsymbol{x}_{i},\boldsymbol{x}_{j})\forall i,j\in{1,\ldots,t}$,
}and
\begin{center}
$\mu_{t+1}(\boldsymbol{x})=\boldsymbol{k}^{T}[K+\sigma_{noise}^{2}I]^{-1}y_{1:t}$
\qquad{}\qquad{}$\sigma_{t+1}(\boldsymbol{x})=k(\boldsymbol{x},\boldsymbol{x})-\boldsymbol{k}^{T}[K+\sigma_{noise}^{2}I]^{-1}\boldsymbol{k}$
\par\end{center}

Next, a surrogate utility function called \emph{acquisition function
}is constructed to find the next sample to evaluate. It balances two
contrastive needs of sampling at the high mean location versus sampling
at the high uncertainty location so that the global optima for {\small{}$f(.)$}
is reached in a fewer number of samples. Acquisition functions are
either constructed based on improvement over the current best (e.g.
Probability of Improvement, Expected Improvement \cite{kushner1964new,mockus1978toward}),
or information based criteria (e.g. Entropy Search \cite{hennig2012entropy}.
Predictive Entropy Search \cite{hernandez2014predictive}), or confidence
based criteria (e.g. GP-UCB \cite{Srinivas_etal_2010Gaussian}). A
GP-UCB acquisition function for the {\small{}$(t+1)$} th iteration
is: 
\begin{center}
$\text{a}_{t+1}(\boldsymbol{x})=\mu_{t}(\boldsymbol{x})+\sqrt{\beta_{t+1}}\sigma_{t}(\boldsymbol{x})$
\par\end{center}

where {\small{}$\beta_{t}$} is an increasing sequence of {\small{}$O(log\thinspace t)$}.
An example sequence can be {\small{}$\beta_{t}=2log(t^{d/2+2}\pi^{2}/3\delta)$}
where {\small{}$d$} represents the dimensionality of the data and
{\small{}$1-\delta$} is the probability of convergence. A regret
$r_{t}$ is defined as the difference of the {\small{}$t$}'th function
evaluation and the global maxima, i.e. {\small{}$r_{t}=max_{\boldsymbol{x}\in\mathcal{X}}f(\boldsymbol{x})-f(\boldsymbol{x}_{t})$},
and the cumulative regret is defined as {\small{}$R_{t}=\sum_{t'=1}^{t}r_{t'}$}.
It can be shown that when Gaussian process is used with SE kernel
then {\small{}$R_{t}\sim O(\sqrt{t(log\thinspace t)^{d+1}})$}, \emph{i.e}.
it only grows sublinearly and{\small{} $Lim_{t\rightarrow\infty}R_{t}/t\rightarrow0$},
implying a `no regret' algorithm. A generic Bayesian optimisation
is a sequential algorithm with one recommendation per iteration. However,
when at each iteration it is convenient to perform a batch of recommendation.
it can be altered to produce a batch of recommendations at each iteration.
Some of the popular batch Bayesian optimisation algorithms with theoretical
guarantee include BUCB \cite{desautels2014parallelizing}, and GP-UCB-PE
\cite{contal_2013parallel}.

\section{Proposed framework\label{sec:framework}}

As previously mentioned, we model the control function $g(t)$ using
the Bernstein polynomial basis. Instead of optimising the function
$g(t)$ directly, we optimise its Bernstein coefficients $\{\alpha_{v}\}_{v=1:n}$.
In this manner, we are able to convert our functional optimisation
problem into a vector optimisation problem. In this section, first
we present how the optimum control function can be found in the presence
of basic shape information. We also discuss how the order of the polynomial
can be adjusted based on the complexity of the control function being
optimised.

\subsection{{\normalsize{}Bernstein polynomial representation with shape constraints}}

An $n^{th}$ order Bernstein polynomial as a linear combination of
its basis polynomials is represented as

\qquad{}\qquad{}\qquad{}$g_{n}(t)=\sum_{v=0}^{n}\alpha_{v}b_{v,n}(t)$\hfill{}(1)

\noindent where $b_{v,n}(t)=\binom{n}{v}t^{v}(1-t)^{n-v}$ are the
Bernstein basis polynomials for order $n$ defined on $[0,1]$ and
$\binom{n}{v}$ is the binomial coefficient, and $\alpha_{v}$ are
the Bernstein coefficients. In other words, the Bernstein polynomial
is the weighted sum of the basis polynomials. We first present a lemma
that guarantees universality of Bernstein polynomial basis.

\textbf{Lemma 1 }\cite{bernstein1912demonstration}\textbf{.}\textbf{\emph{
}}\emph{Any continuous function $f$ defined on the closed interval
$[0,1]$ can be uniformly approximated by a Bernstein polynomial function
$B_{n}(f)$. Let $B_{n}(f)(t)=\sum_{v=0}^{n}f\left(\frac{v}{n}\right)b_{v,n}(t)$.
then as $n\to\infty$, $B_{n}(f)$ converges to the function $f$,
i.e.} $lim_{n\to\infty}B_{n}(f)=f$. \hfill{}$\oblong$

Next, we present lemmas to control the shape of the function. Our
interest lies in the elegant relationship between the Bernstein coefficients
$\{\alpha_{v}\}_{v=0}^{n}$ and the shape of the Bernstein polynomial.
In Theorem 1 and 2, we will elaborate on the details of this relationship
for the monotonic function and the unimodal function case. The following
Lemma leads us towards the statements in these theorems. 

\textbf{Lemma 2. }\emph{For a Bernstein polynomial $g_{n}(t)=\sum_{b=0}^{n}\alpha_{v}b_{v,n}(t)$,
the derivative of the polynomial is given by $g_{n}^{'}(t)=n\sum_{v=0}^{n-1}\left(\alpha_{v+1}-\alpha_{v}\right)b_{v,n-1}(t)$.
In other words the derivative of the $n$th order Bernstein polynomial
can be expressed through a linear combination of Bernstein base polynomials
up to order $(n-1)$.} (Please refer to supplementary material detailed
proof)\hfill{}$\oblong$

\textbf{Theorem 1. }(Monotonicity)\cite{chang2007shape}\textbf{:}\emph{
If $\alpha_{v+1}\geq\alpha_{v}$, then $g_{n}(t)$ is a monotonically
increasing function. Similarly if $\alpha_{v+1}\leq\alpha_{v}$, then
$g_{n}(t)$ is a monotonically decreasing function. }

\emph{Proof:}\textbf{\emph{ }}From Lemma 2, consider the derivative
of the Bernstein polynomial \emph{$g_{n}^{'}(t)=n\sum_{v=0}^{n-1}\left(\alpha_{v+1}-\alpha_{v}\right)b_{v,n-1}(t)$.}
Here, the base polynomials $\left\{ b_{v,n-1}(t)\right\} _{v=0}^{n-1}$
are always positive by definition. Therefore, if the difference $\left(\alpha_{v+1}-\alpha_{v}\right)$
is kept positive, the derivative $g_{n}^{'}(t)$ remains positive
implying \emph{$g_{n}(t)$} to be a monotonically increasing function.
Similar argument can be made for $g_{n}(x)$ to be a monotonically
decreasing function provided $\alpha_{v+1}<\alpha_{v}$. \hfill{}$\oblong$

\textbf{Theorem 2} (Unimodality)\cite{chang2007shape}\textbf{:}\textbf{\emph{
}}\emph{For $n\geq3$. If $\alpha_{0}=\alpha_{1}=...=\alpha_{l_{1}}<\alpha_{l_{1}+1}\leq\alpha_{l_{1}+2}\leq...\leq\alpha_{l_{2}}$
and $\alpha_{l_{2}}\geq\alpha_{l_{2}+1}\geq...\geq\alpha_{l_{3}}>\alpha_{l_{3}+1}=...=\alpha_{n}$
for some $0\leq l_{1}<l_{2}\leq l_{3}\leq n$, then there exists some
$s\in(0,\tau)]$ such that $s$ is the unique maximum point of $g_{n}(t)$
and $g_{n}(t)$ is strictly increasing on $[0,s]$ and it is strictly
decreasing on $[s,\tau].$}

\emph{Proof:}\textbf{\emph{ }}Please refer to \cite{chang2007shape}
for proof. The proof also uses Lemma 2 as a key ingredient.  \hfill{}$\oblong$

These theorems are used to formulate constraints in the next subsection.
 Chang et al. \cite{chang2007shape} have also provided theory for
the cases other than unimodal concave. Our framework can be extended
to all such cases where such a relationship between the coefficients
and the shape of the polynomial has been established.\emph{ }Often
the prior information about the trend is available from the domain
experts. Specifically, in the design of short polymer fibre, it is
known that as the fibre recirculates they get shorter. Hence, to get
more narrow size distribution we need to produce longer fibre initially,
with gradual reduction in length. This would require butanol flow
to increase monotoically with time. Similarly, for deep learning learning
schedule tuning we also use monotonically decreasing constraint. However,
our method can be applied to applications where the function is either
unimodal or even when there is no prior information available about
the shape of the control function. 

\begin{figure}
\begin{centering}
\subfloat[Bernstein base polynomials for the third order Bernstein polynomial.]{\centering{}\includegraphics[width=1\columnwidth]{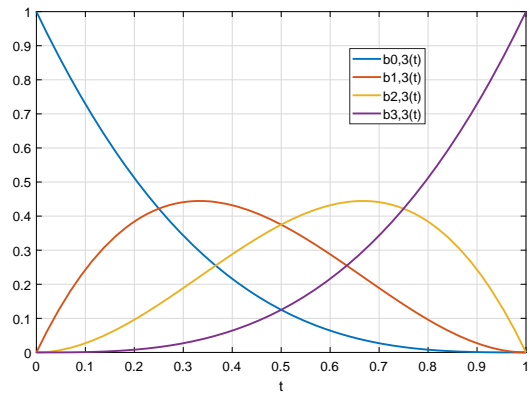}}
\par\end{centering}
\centering{}\subfloat[Examples of Bernstein polynomial $g_{3}(t)$ based on how alphas are
sampled.]{\begin{centering}
\includegraphics[width=1\columnwidth]{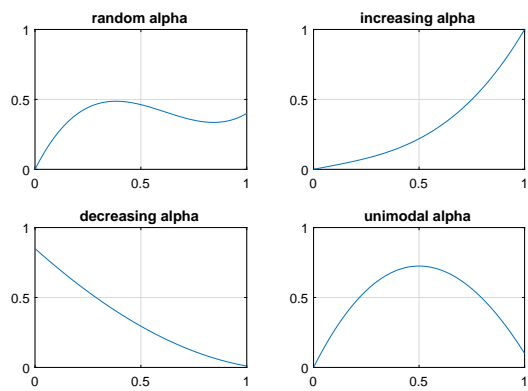}
\par\end{centering}
}\caption{{\small{}Examples of a third order Bernstein polynomial. \label{fig:Bpoly}}}
\end{figure}

For illustration, let us consider the functions that are represented
via third order Bernstein polynomial basis. The base polynomials in
this case are $b_{0,3}(t)=(1-t)^{3}$, $b_{1,3}(t)=3t(1-t)^{2}$,
$b(t)=3t^{2}(1-t),$ and $b_{3,3}(t)=t^{3}$. Fig \ref{fig:Bpoly}
shows the sample functions from functions with only range constraint,
as well as functions with monotonicty constraint and unimodality constraint.

\begin{algorithm*}
\textbf{\small{}Input: }{\small{}Observations $\mathcal{D}_{1:m}=\{\boldsymbol{x}_{i},y_{i}\}_{i=1}^{m}$,
where $y_{i}=f(g_{n}(t)\given\boldsymbol{x}_{i})+\varepsilon$ for
an $n$th order Bernstein polynomial, $\boldsymbol{x}_{i}=\{\boldsymbol{\alpha}_{i},\boldsymbol{u}_{i}\}$
for $\boldsymbol{\alpha}_{i}\in\mathbb{R}^{1\times n}$, Fixed increment
schedule $\omega$.}{\small\par}

\textbf{\small{}for $m=1:MaxIteration$ }{\small\par}

\begin{onehalfspace}
{\small{}\qquad{}build $\mathcal{GP}$ on $\mathcal{D}_{1:m}$}{\small\par}

{\small{}\qquad{}sample $\boldsymbol{x}_{m+1}=\text{argmax}_{x\in\mathcal{X}}\text{a}\left(\boldsymbol{x}_{m}\given C,\mathcal{D}\right)$}{\small\par}

{\small{}\qquad{}evaluate $y_{m+1}=f(g_{n}(t)\given\boldsymbol{x}_{m+1})$
(see eq: (1))}{\small\par}

{\small{}\qquad{}obtain the current best sample $\{\boldsymbol{x}^{+},y^{+}\}$}{\small\par}

{\small{}\qquad{}compute the maximum difference $d$ between any
two coefficients in $\boldsymbol{\alpha}^{+}$}{\small\par}

{\small{}\qquad{}}\textbf{\small{}if}{\small{} $d>0.95$ or $m\%\omega=0$}{\small\par}

{\small{}\qquad{}\qquad{}update order $n=n+1$ and re-evaluate $\{\boldsymbol{\alpha}_{i}\}_{i=1}^{m}$
for $\boldsymbol{\alpha}_{i}\in\mathbb{R}^{1\times(n+1)}$}{\small\par}

{\small{}\qquad{}\qquad{}update observations $\mathcal{D}_{1:m}$}{\small\par}

{\small{}\qquad{}}\textbf{\small{}end if}{\small\par}

{\small{}\qquad{}augment the data $\mathcal{D}_{1:m+1}=\mathcal{D}_{1:m}\cup\left\{ \boldsymbol{x}_{m+1},y_{m+1}\right\} $}{\small\par}
\end{onehalfspace}

\textbf{\small{}end for }{\small\par}

{\small{}\caption{Framework for control function optimisation{\small{}. \label{alg:schedule_with_trend} }}
}{\small\par}
\end{algorithm*}

\subsection{{\normalsize{}Control  function optimisation with shape constraints}}

We recall our optimisation problem as:
\begin{center}
$\boldsymbol{x}^{*}=\text{argmax}_{x\in\mathcal{X}}f(\boldsymbol{x})$
\par\end{center}

\noindent where $\boldsymbol{x}=\{\boldsymbol{\alpha},\boldsymbol{u}\}$,
$\mathcal{X}=\mathcal{A}\cup\mathcal{U}$, $\boldsymbol{\alpha}$
is the Bernstein coefficient vector and $\boldsymbol{u}$ are the
other fixed control variables. When prior information about the shape
of the control  function is available, then it boils down to usual
constrained optimisation problem as:
\begin{center}
$\boldsymbol{x}^{*}=\text{argmax}_{x\in\mathcal{X}}f(\boldsymbol{x})$\qquad{}\qquad{}$s.t.$
$\mathbb{C}\geq0$
\par\end{center}

\noindent where $\mathbb{C}$ is a set of inequality constraints over
the parameter space. Such constraints can be easily enforced during
acquisition function optimisation. In the following we present the
$\mathbb{C}$ for monotonically increasing and decreasing control
 functions,  and unimodal control  functions based on the Theorems
1 and 2:

\paragraph*{\emph{Controlling the range of the control function:}}

By choosing the values of $\alpha_{0:n}\in[0,1]$ the Bernstein polynomial
is limited to the range $[0,1]$ (this is possible due to Lemma 3,
which will be described in the next subsection). For any application,
optimised Bernstein polynomial can then be rescaled using the known
range of input function.

\paragraph*{\emph{Increasing control  function:}}

$\mathbb{C}=\{C_{v}\}_{v=0}^{n-1}$ where $C_{v}=$$\alpha_{v+1}-\alpha_{v}$.
This is used in the design of recirculation control  function in short
polymer fibre production experiment as presented in experiments.

\paragraph*{\emph{Decreasing control  function:} }

$\mathbb{C}=\{C_{v}\}_{v=0}^{n-1}$ where $C_{v}=$$\alpha_{v}-\alpha_{v+1}$.
These constraints are used in modelling learning rate schedules for
deep neural network optimisers as presented in experiments.

\paragraph*{\emph{Unimodal control  function:} }

For unimodal control function we chose simplified constraints adapted
from Theorem 2. For some $0<l<n$, we define the constraints as $\mathbb{C}=\{C_{v}\}_{v=0}^{n-1}$
where $C_{v}=$ $\alpha_{v+1}-\alpha_{v}$ $\forall v=0:l$ and $C_{v}=$
$\alpha_{v}-\alpha_{v+1}$ $\forall v=l:n-1$.

\subsection{{\normalsize{}Dynamically adjusting order of Bernstein basis}}

As discussed earlier the intuition is illustrated in the figure 2.
We compare the maximum computed derivative of the best control function
with the maximum derivative possible with the same order polynomial.
If it is close then we increase the polynomial order.

We first derive Lemma 3 that provides an easy way to compute the maximum
of the derivative of a function based on the Bernstein coefficients. 

\textbf{Lemma 3.}\emph{ The derivative $g_{n}^{'}(t)$ of the Bernstein
polynomial $g_{n}(t)$ is bounded by `n' times the maxima of the basis
polynomial with the highest coefficient, where `n' is the order.}

\emph{Proof:} It can be said about Bernstein polynomials that the
range of the polynomial $g_{n}(t)$ is bounded by the values of the
minimum and maximum values of the Bernstein coefficients $\{\alpha_{v}\}_{v=0:n}$.
Then from Lemma 2 we can say that the derivative of the Bernstein
polynomial \emph{$g_{n}^{'}(t)$} is bounded by the values of the
minimum and maximum of the values of the Bernstein coefficients $\{n(\alpha_{v+1}-\alpha_{v})\}_{v=0:n-1}$.
By the same token the theoretical maximum of the magnitude of the
derivative is $(\alpha_{max}-\alpha_{min})$, assuming $\alpha\in[\alpha_{max},\thinspace\alpha_{min}]$.
\hfill{}$\oblong$

Based on this it can be said that an $n$-th order Bernstein polynomial
can only be used to sufficiently approximate a function if the derivate
of the function is within the bounds as stated in Lemma 3. We should
test whether the current function has already reached the limit or
at least close to it. If it is true then we can decide that the current
order is underspecified and we increment the order by one. 

Next, we use the following lemma to reuse the past observations by
transforming $\{\alpha_{v}\}_{v=0:n}$ vectors of length $n$ to $\{\alpha_{v}^{'}\}_{v=0:n+1}$
vectors of length $n+1$.

\textbf{Lemma 4.}\emph{ A Bernstein polynomial of order $n$ and with
Bernstein coefficients} $\{\alpha_{v}\}_{v=0:n}$\emph{ can be represented
using a Bernstein polynomial of order $n+1$ with Bernstein coefficients}
$\{\alpha_{v}^{'}\}_{v=0:n+1}$\emph{, such that $\alpha_{v}^{'}=\frac{v}{n+1}\alpha_{v-1}+\left(1-\frac{v}{n+1}\right)\alpha_{v}$.
In other words, it is possible to raise the order of the Bernstein
polynomial and recompute the Bernstein coefficients.}

\emph{Proof: }Please refer to \cite{lorentz1953bernstein} . \hfill{}$\oblong$

Lemma 3 can detect one type of signs of underspecification. To avoid
underspecification altogether, we also increment the order at a regular
interval until a maximum specified order is reached. The overall algorithm
is presented in Algo \ref{alg:schedule_with_trend}.

\section{Experiments\label{sec:Experiments}}

We evaluate our proposed functional optimisation method on one synthetic
and two real world experiments: optimisation of fibre yield in short
polymer fibre production, and learning rate schedule optimisation
for neural network training. For convenience, we refer to the proposed
algorithm as BFO-SP. For all experiments we start with a 5th order
Bernstein polynomial basis, but limiting to 10 as the highest order.
The change of order is triggered due to hitting the derivative limit
when it reaches 95\% of the maximum derivative magnitude possible.
The code will be made available at \emph{https://goo.gl/6twf4i}.

\begin{figure}
\subfloat[With monotonicity constraint.]{\begin{centering}
\includegraphics[width=0.9\columnwidth]{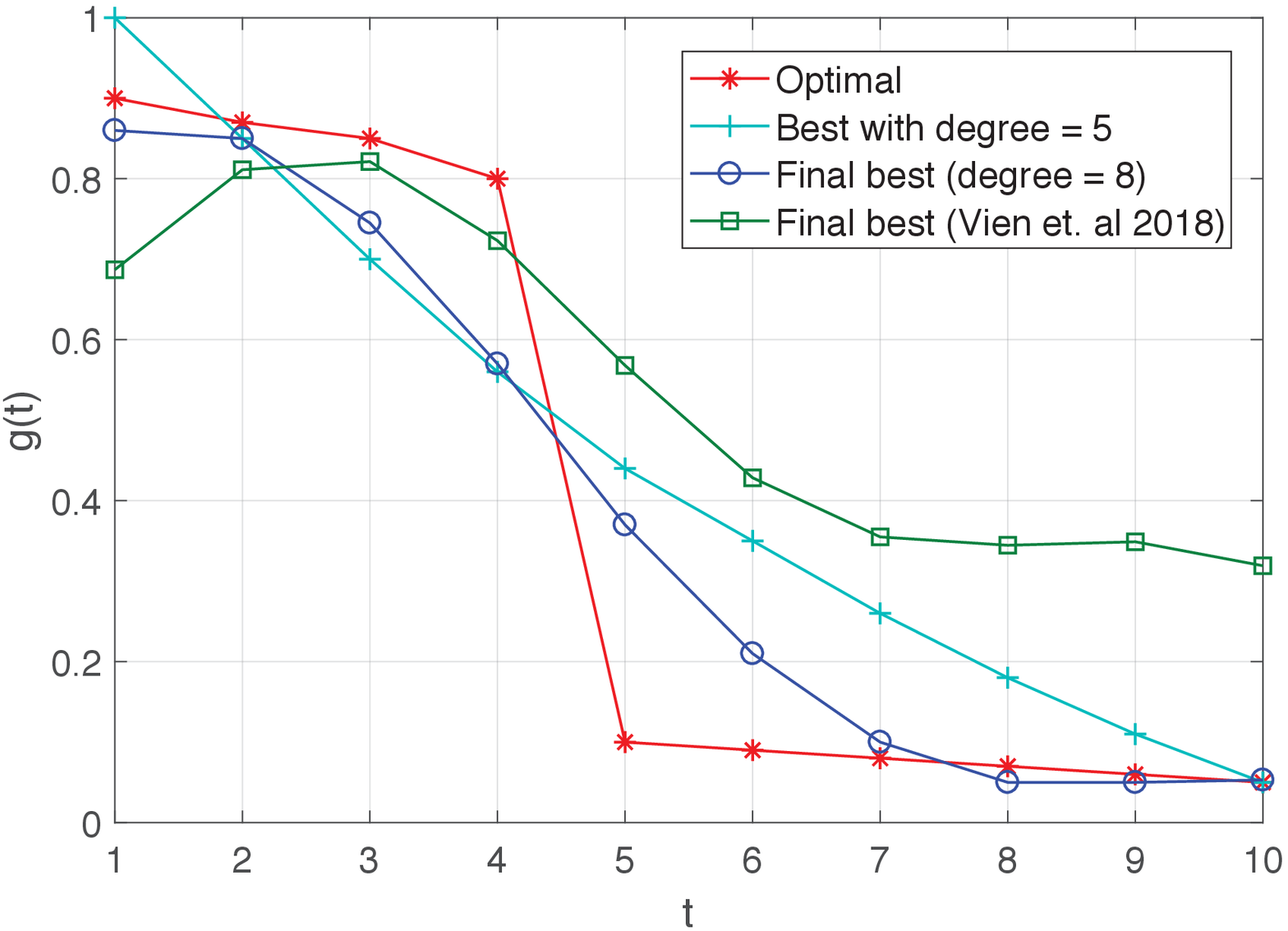}
\par\end{centering}
}

\subfloat[With unimodality constraint.]{\begin{centering}
\includegraphics[width=0.9\columnwidth]{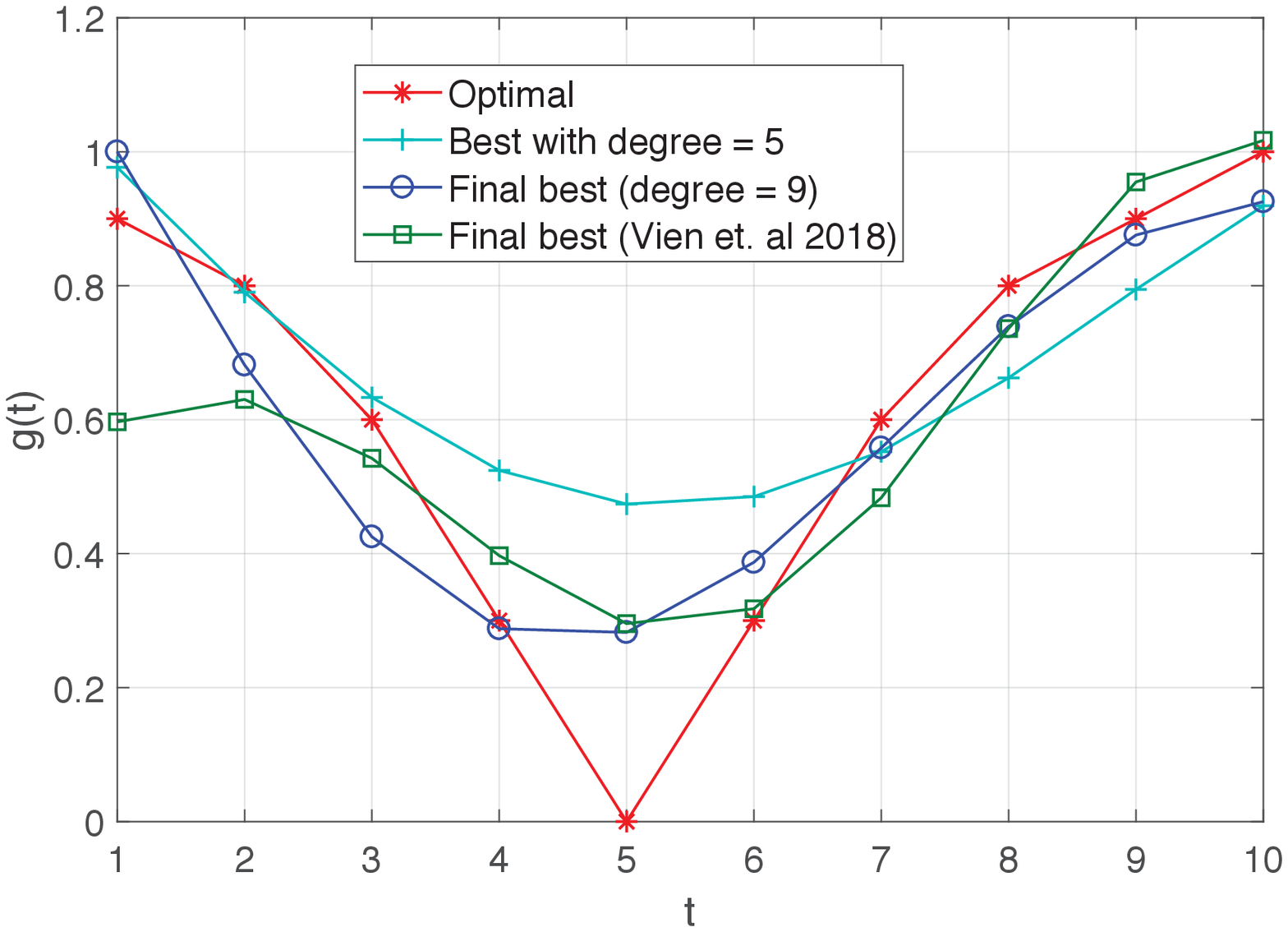}
\par\end{centering}
}

\caption{{\small{}Synthetic experiment.\label{fig:Synthetic-experimen} }}

\noindent\begin{minipage}[t]{1\columnwidth}%
\begin{center}
\includegraphics[width=1\columnwidth]{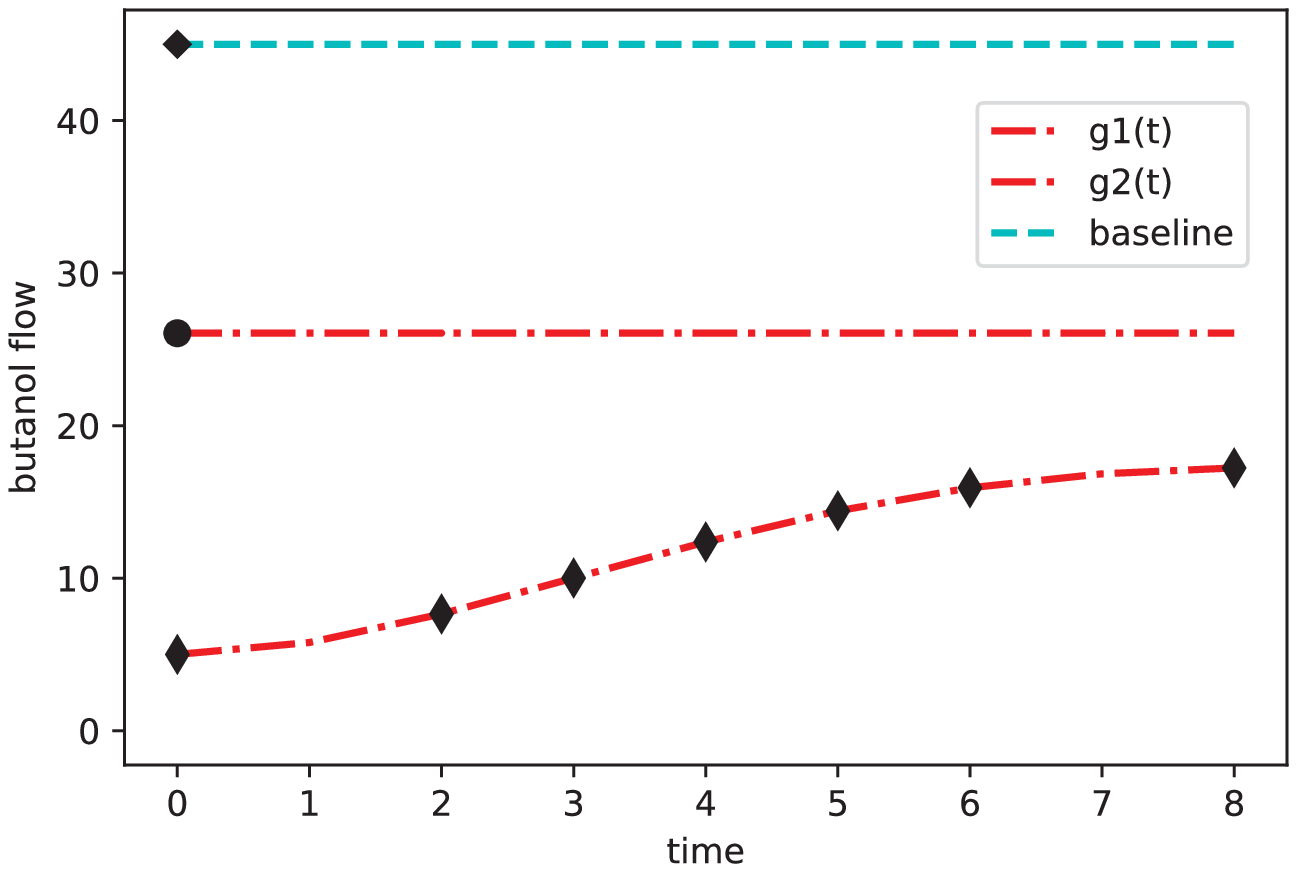}
\par\end{center}
\caption{{\small{}Short polymer fibre production: Best recirculation control
functions $g(t)$ found compared to baseline. $g1(t)$ and $g2(t)$
gave highest score (9). \label{fig:recirc_sched}}}
\end{minipage}
\end{figure}

\subsection{Synthetic experiments }

In the synthetic experiment we construct two different 10 dimensional
vectors, one which is monotonically decreasing and the other which
is unimodal convex. These can be thought of schedule vectors, sampled
on a fixed grid of size 10 from our optimal control functions that
we would like to recover via our functional optimisation approach.
For each, the utility of a trial control function is measured by first
sampling the control function on the same grid and evaluating the
resultant 10 dimensional vector through a Gaussian pdf function whose
mean is the respective optimal vector. We start with Bernstein polynomial
of order 5 and then increase the order when either a) at the trigger
as mentioned in lemma 2, or b) at a fixed schedule of every 10 iterations.
The results are shown in Fig \ref{fig:Synthetic-experimen} with a)
showing the case for monotonicity and b) showing the case for unimodality.
In the plots we show the optimal (in red), the best so far when the
first order change is triggered by lemma 2 (in teal) and the best
one after 20 iterations. For both the cases order changes have been
triggered multiple times resulting in the final best with considerable
higher order (8 for monotonicity and 9 for unimodality). The final
best ones looks much closer to the optimal one. The results are compared
with the method proposed by \cite{vien2018bayesian} (in green), where
incorporating prior is not possible. 

\begin{table*}
\begin{centering}
{\small{}}%
\begin{tabular}{|c||c|c|c|c|}
\hline 
\multirow{2}{*}{\textbf{\small{}Dataset}} & \multicolumn{4}{c|}{\textbf{\small{}Validation error}}\tabularnewline
\cline{2-5} 
 & \textbf{\small{}BFO-SP + SGD } & \textbf{\small{}SGD } & \textbf{\small{}Adam} & \textbf{\small{}BFO + SGD (Vien et al. 2018)}\tabularnewline
\hline 
\hline 
\textbf{\small{}CFIR10} & \textbf{\small{}18.81\% } & {\small{}20.30\%} & {\small{}20.20\%} & {\small{}22.2\%}\tabularnewline
\hline 
\textbf{\small{}MNIST} & \textbf{\small{}0.74\% } & {\small{}1.26\%} & {\small{}0.86\%} & {\small{}0.87\%}\tabularnewline
\hline 
\end{tabular}\vspace{0.005\textheight}
\par\end{centering}
\caption{{\small{}Comparison of prediction error of Bayesian optimisation of
learning rate schedule against SGD and Adam with exponential decay
for both CFIR10 and MNIST datasets. \label{tab:Comparison}}}
\end{table*}

\subsection{Short polymer fibre production}

We optimise the butanol flow profile over the recirculation period
to achieve a high quality yield of concentrated fibres. All other
variables (device geometry and polymer flow) are fixed to the known
best setting. The recirculation is run till a fixed time, limited
by the maximum concentration achievable with the chosen polymer flow.
At the end of each experiment a sample of fibre is looked under a
powerful optical microscope to inspect fibre length and diameter distribution.
Quality score is given between 1-10, with 10 being the highest for
fibre distribution with small variance. Experimenters had a hunch
that an increasing flow profile will result in a higher quality yield,
which we used as our shape prior. We used GP-UCB-PE with batch size
of 6.

We have been able to reach a score of 9 out of a maximum score of
10, within 5 iterations. Figure \ref{fig:recirc_sched} shows examples
of butanol flow schedules for which high scores were recorded. Both
a flat and increasing profile results in a score of 9, thus validating
the experimenters hunch. These are improvements over their current
baseline with a fixed butanol flow (8). The markers on each function
are time intervals at which the butanol flow is{\small{} changed.}

\begin{figure}[H]
\noindent\begin{minipage}[t]{1\columnwidth}%
\begin{center}
\includegraphics[width=1\columnwidth]{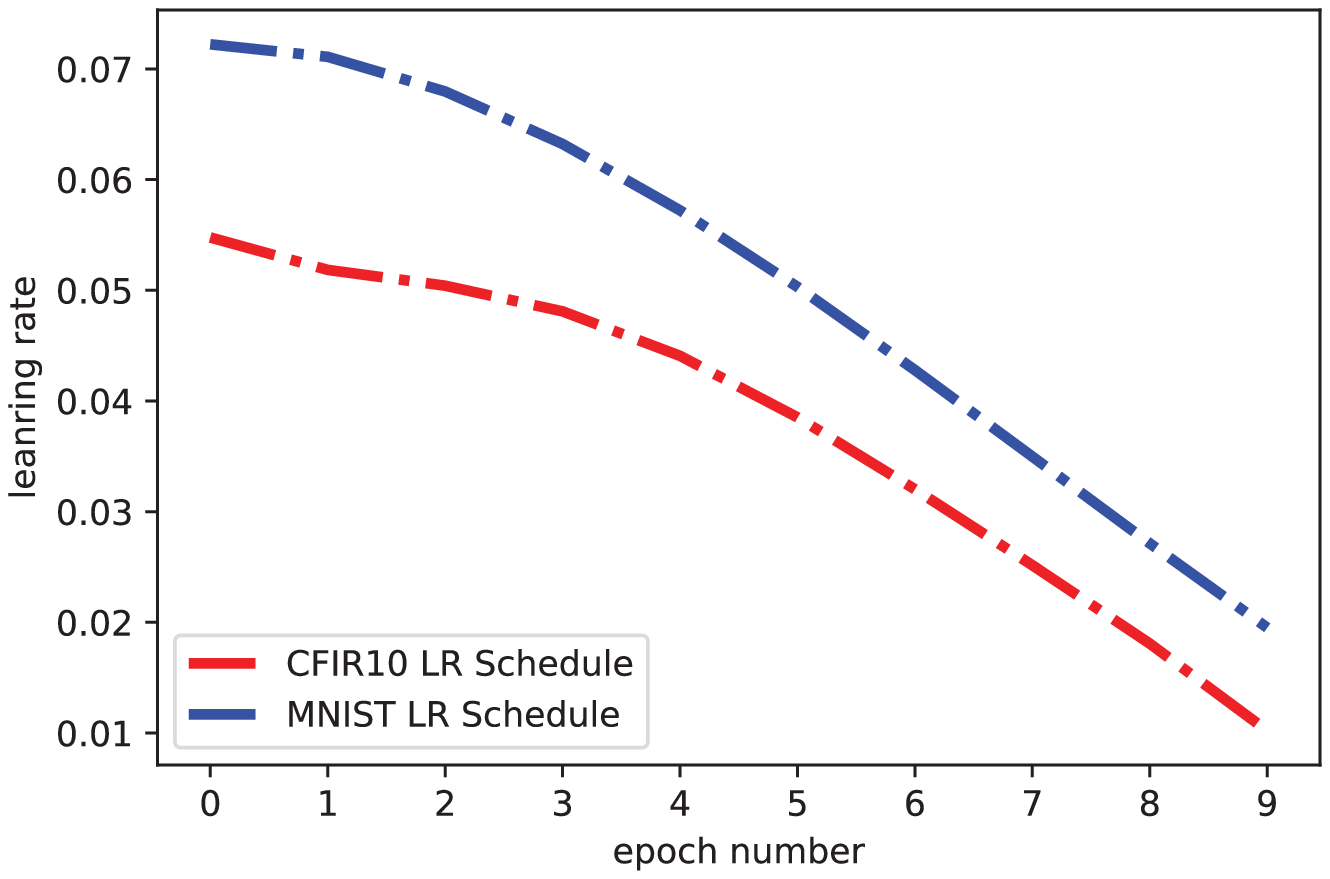}
\par\end{center}
\caption{{\small{}Learning rate schedules that resulted in the highest accuracy
on CFIR10 and MNIST datasets using the BFO-SP with known prior - monotonicity
constraint.\label{fig:lr_func}}}
\end{minipage}

\noindent\begin{minipage}[t]{1\columnwidth}%
\begin{center}
\includegraphics[width=1\columnwidth]{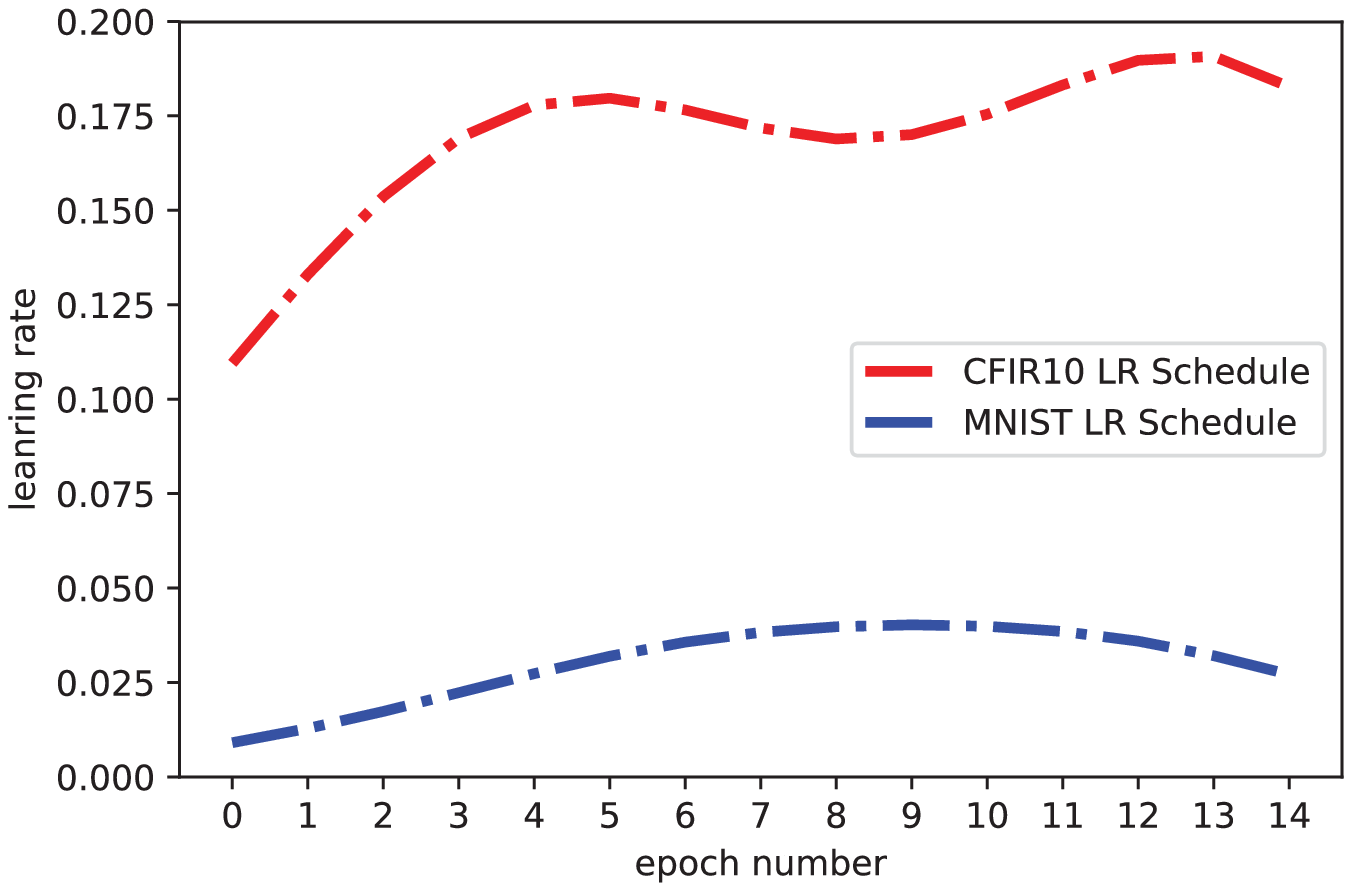}
\par\end{center}
\caption{{\small{}Learning rate schedules that resulted in the highest accuracy
on CFIR10 and MNIST datasets using \cite{vien2018bayesian}\label{fig:vien}}}
\end{minipage}
\end{figure}

\subsection{Learning rate schedule optimisation}

For neural network training it has been observed that stochastic gradient
descent (SGD) performs better if the learning rate is varied as a
function of training duration \cite{bengio2012practical}. Specifically,
it has been reported that starting with an adequately high learning
rate and decreasing it over the training duration can significantly
speed-up convergence. We optimise for a schedule of learning rate
$f(\eta)$ for a couple of neural networks, each for the CFIR10 and
MNIST datasets, while keeping the other parameters the same for all
the experiments.

For CFIR10 we use a network architecture that be summarised as $(Conv2D\to Dropout\to Conv2D\to Maxpooling2D)\times3\to Flatten\to(Dropout\to Dense)\times3$.
Whereas for MNIST the network architecture used is $(Conv2D\to Maxpooling2D\to Dropout\to Flatten\to Dense\to Dense)$.
(Details about the network architecture are given in the supplementary
material). Fig \ref{fig:lr_func} shows the optimised learning schedules
for CFIR10 and MNIST datasets. For Bayesian optimisation the range
of learning rate was chosen between 0.2 and 0.0001. For Adam and SGD
the starting learning rate used was 0.01, with 0.8 momentum for SGD
and default values for hyper-parameters of Adam \cite{kingma2014adam}.
We compare the performance of learning rate optimised stochastic gradient
descent (SGD) optimiser against a) SGD with an exponential decay and
b) Adam. Table \ref{tab:Comparison} shows the performance comparison
with the baselines after 20 Bayesian optimisation iteration. On both
the datasets our method BFO-SP achieved higher performance than the
baselines. The results are also compared with BFO by \cite{vien2018bayesian}
as shown in Fig \ref{alg:schedule_with_trend}. By admitting prior
knowledge, in limited number of iterations our method seems to perform
well.

\section{Conclusion}

We present a novel approach for functional optimisation with Bayesian
optimisation. We use Bernstein polynomials to model the control function
and in turn optimise the Bernstein coefficients to learn the optimum
function shape. Prior shape information (e.g. monotonicity, unimodality
etc) is integrated and the polynomial order is dynamically adjusted
during the optimisation process. We demonstrate the performance of
our method by applying it for short polymer fibre recirculation production,
and for modelling learning rate schedule for deep learning networks.
Our method performs well in both the cases and can be useful in many
industrial processes involving recirculation.

\newpage{}

\bibliographystyle{aaai}
\bibliography{Vellanki_BO}

\begin{thebibliography}{}

\bibitem[\protect\citeauthoryear{Bengio}{2012}]{bengio2012practical}
Bengio, Y.
\newblock 2012.
\newblock Practical recommendations for gradient-based training of deep
  architectures.
\newblock In {\em Neural networks: Tricks of the trade}. Springer.
\newblock  437--478.

\bibitem[\protect\citeauthoryear{Bernstein}{1912}]{bernstein1912demonstration}
Bernstein, S.
\newblock 1912.
\newblock Demonstration of a theorem of weierstrass based on the calculus of
  probabilities.
\newblock {\em Communications of the Kharkov Mathematical Society} Volume
  XIII(1912/13):1--2.

\bibitem[\protect\citeauthoryear{Brochu, Cora, and {de
  Freitas}}{2010}]{Brochu_Cora_Freitas_2010Tutorial}
Brochu, E.; Cora, V.~M.; and {de Freitas}, N.
\newblock 2010.
\newblock A tutorial on {B}ayesian optimization of expensive cost functions,
  with application to active user modeling and hierarchical reinforcement
  learning.
\newblock {\em arXiv:1012.2599} (UBC TR-2009-023 and arXiv:1012.2599).

\bibitem[\protect\citeauthoryear{Chang \bgroup et al\mbox.\egroup
  }{2007}]{chang2007shape}
Chang, I.-S.; Chien, L.-C.; Hsiung, C.~A.; Wen, C.-C.; and Wu, Y.-J.
\newblock 2007.
\newblock Shape restricted regression with random bernstein polynomials.
\newblock {\em Lecture Notes-Monograph Series}  187--202.

\bibitem[\protect\citeauthoryear{Contal \bgroup et al\mbox.\egroup
  }{2013}]{contal_2013parallel}
Contal, E.; Buffoni, D.; Robicquet, A.; and Vayatis, N.
\newblock 2013.
\newblock Parallel gaussian process optimization with upper confidence bound
  and pure exploration.
\newblock In {\em Joint European Conference on Machine Learning and Knowledge
  Discovery in Databases},  225--240.
\newblock Springer.

\bibitem[\protect\citeauthoryear{Desautels, Krause, and
  Burdick}{2014}]{desautels2014parallelizing}
Desautels, T.; Krause, A.; and Burdick, J.~W.
\newblock 2014.
\newblock Parallelizing exploration-exploitation tradeoffs in gaussian process
  bandit optimization.
\newblock {\em Journal of Machine Learning Research} 15(1):3873--3923.

\bibitem[\protect\citeauthoryear{Feng \bgroup et al\mbox.\egroup
  }{2003}]{feng2003creation}
Feng, L.; Song, Y.; Zhai, J.; Liu, B.; Xu, J.; Jiang, L.; and Zhu, D.
\newblock 2003.
\newblock Creation of a superhydrophobic surface from an amphiphilic polymer.
\newblock {\em Angewandte Chemie International Edition} 42(7):800--802.

\bibitem[\protect\citeauthoryear{Hennig and Schuler}{2012}]{hennig2012entropy}
Hennig, P., and Schuler, C.~J.
\newblock 2012.
\newblock Entropy search for information-efficient global optimization.
\newblock {\em Journal of Machine Learning Research} 13(Jun):1809--1837.

\bibitem[\protect\citeauthoryear{Hern{\'a}ndez-Lobato, Hoffman, and
  Ghahramani}{2014}]{hernandez2014predictive}
Hern{\'a}ndez-Lobato, J.~M.; Hoffman, M.~W.; and Ghahramani, Z.
\newblock 2014.
\newblock Predictive entropy search for efficient global optimization of
  black-box functions.
\newblock In {\em Advances in neural information processing systems},
  918--926.

\bibitem[\protect\citeauthoryear{Jones, Schonlau, and
  Welch}{1998}]{jones1998efficient}
Jones, D.~R.; Schonlau, M.; and Welch, W.~J.
\newblock 1998.
\newblock Efficient global optimization of expensive black-box functions.
\newblock {\em Journal of Global optimization} 13(4):455--492.

\bibitem[\protect\citeauthoryear{Kingma and Ba}{2014}]{kingma2014adam}
Kingma, D.~P., and Ba, J.
\newblock 2014.
\newblock Adam: A method for stochastic optimization.
\newblock {\em arXiv preprint arXiv:1412.6980}.

\bibitem[\protect\citeauthoryear{Kulfan and
  Bussoletti}{2006}]{kulfan2006fundamental}
Kulfan, B., and Bussoletti, J.
\newblock 2006.
\newblock " fundamental" parameteric geometry representations for aircraft
  component shapes.
\newblock In {\em 11th AIAA/ISSMO multidisciplinary analysis and optimization
  conference},  6948.

\bibitem[\protect\citeauthoryear{Kushner}{1964}]{kushner1964new}
Kushner, H.~J.
\newblock 1964.
\newblock A new method of locating the maximum point of an arbitrary multipeak
  curve in the presence of noise.
\newblock {\em Journal of Basic Engineering} 86(1):97--106.

\bibitem[\protect\citeauthoryear{Lorentz}{1953}]{lorentz1953bernstein}
Lorentz, G.~G.
\newblock 1953.
\newblock {\em Bernstein polynomials}.
\newblock American Mathematical Soc.

\bibitem[\protect\citeauthoryear{Ma, Hill, and Rutledge}{2008}]{ma2008review}
Ma, M.; Hill, R.~M.; and Rutledge, G.~C.
\newblock 2008.
\newblock A review of recent results on superhydrophobic materials based on
  micro-and nanofibers.
\newblock {\em Journal of Adhesion Science and Technology} 22(15):1799--1817.

\bibitem[\protect\citeauthoryear{Mockus, Tiesis, and
  Zilinskas}{1978}]{mockus1978toward}
Mockus, J.; Tiesis, V.; and Zilinskas, A.
\newblock 1978.
\newblock Toward global optimization, volume 2, chapter bayesian methods for
  seeking the extremum.

\bibitem[\protect\citeauthoryear{Mockus}{1994}]{mockus1994application}
Mockus, J.
\newblock 1994.
\newblock Application of bayesian approach to numerical methods of global and
  stochastic optimization.
\newblock {\em Journal of Global Optimization} 4(4):347--365.

\bibitem[\protect\citeauthoryear{Rasmussen}{2006}]{rasmussen2006gaussian}
Rasmussen, C.~E.
\newblock 2006.
\newblock Gaussian processes for machine learning.

\bibitem[\protect\citeauthoryear{Samareh}{2001}]{samareh2001survey}
Samareh, J.~A.
\newblock 2001.
\newblock Survey of shape parameterization techniques for high-fidelity
  multidisciplinary shape optimization.
\newblock {\em AIAA journal} 39(5):877--884.

\bibitem[\protect\citeauthoryear{Srinivas \bgroup et al\mbox.\egroup
  }{2010}]{Srinivas_etal_2010Gaussian}
Srinivas, N.; Krause, A.; Kakade, S.; and Seeger, M.~W.
\newblock 2010.
\newblock Gaussian process optimization in the bandit setting: No regret and
  experimental design.
\newblock In {\em International Conference on Machine Learning},  1015--1022.

\bibitem[\protect\citeauthoryear{Vien, Zimmermann, and
  Toussaint}{2018}]{vien2018bayesian}
Vien, N.~A.; Zimmermann, H.; and Toussaint, M.
\newblock 2018.
\newblock Bayesian functional optimization.

\end{thebibliography}

\newpage{}

\section*{Supplementary material}

\selectlanguage{english}%

\subsection{Proof of Lemma 2}

\textbf{Lemma 2. }\emph{For a Bernstein polynomial}\emph{\small{}
$g_{n}(t)=\sum_{b=0}^{n}\alpha_{v}b_{v,n}(t)$}\emph{, the derivative
of the polynomial is given by }\emph{\small{}$g_{n}^{'}(t)=n\sum_{v=0}^{n-1}\left(\alpha_{v+1}-\alpha_{v}\right)b_{v,n-1}(t)$}\emph{.
In other words the derivative of the $n$th order Bernstein polynomial
can be expressed through a linear combination of Bernstein base polynomials
up to order}\emph{\small{} $(n-1)$}\emph{ .} 

\emph{Proof:}\textbf{\emph{ }}The derivative of a Bernstein base polynomial
can be derived as below:
\begin{align*}
b_{v,n}(t) & =\frac{n!}{v!(n-v)!}t^{v}(1-t)^{n-v}\\
\frac{d}{dt}b_{v,n}(t) & =\frac{d}{dt}\left[\frac{n!}{v!(n-v)!}t^{v}(1-t)^{n-v}\right]\\
 & =n\left[\frac{(n-1)!}{(v-1)!(n-v)!}t^{v-1}(1-t)^{n-v}\right]\ldots\\
 & -n\left[\frac{(n-1)!}{v!(n-v-1)!}t^{v}(1-t)^{n-v-1}\right]\\
 & =n\left[b_{v-1,n-1}(t)-b_{v,n-1}(t)\right]
\end{align*}
Now using Eq (1) and above, the derivative of the Bernstein polynomial
$g_{n}(t)$ can be derived as 
\begin{align*}
g_{n}(t) & =\sum_{v=0}^{n}\alpha_{v}b_{v,n}(t)\\
g_{n}^{'}(t) & =\sum_{v=0}^{n}\alpha_{v}n\left[b_{v-1,n-1}(t)-b_{v,n-1}(t)\right]\\
 & \propto\alpha_{0}\left[b_{-1,n-1}(t)-b_{0,n-1}(t)\right]\ldots\\
 & +\alpha_{1}\left[b_{0,n-1}(t)-b_{1,n-1}(t)\right]+\ldots\\
 & +\alpha_{n-1}\left[b_{n-2,n-1}(t)-b_{n-1,n-1}(t)\right]\ldots\\
 & +\alpha_{n}\left[b_{n-1,n-1}(t)-b_{n,n-1}(t)\right]\\
 & =\sum_{v=0}^{n-1}\left(\alpha_{v+1}-\alpha_{v}\right)b_{v,n-1}(t)
\end{align*}
where $b_{-1,n-1}(t)=b_{n,n-1}(t)=0$.

\vspace{0.015\textheight}

\subsection{Network architecture for learning rate schedule (experiment 4.3)}

\begin{table}[H]
\begin{centering}
\subfloat[Network architecture for CIFAR10 experiment.]{\begin{centering}
\begin{tabular}{|c|c|c|}
\hline 
Layer type & Output shape & Param \#\tabularnewline
\hline 
\hline 
Conv2D & (32,32,32) & 896\tabularnewline
\hline 
Dropout & (32,32,32) & 0\tabularnewline
\hline 
Conv2D & (32,32,32) & 9248\tabularnewline
\hline 
MaxPooling2D & (32,16,16) & 0\tabularnewline
\hline 
Conv2D & (64,16,16) & 18496\tabularnewline
\hline 
Dropout & (64,16,16) & 0\tabularnewline
\hline 
Conv2D & (64,16,16) & 36928\tabularnewline
\hline 
MaxPooling2D & (64,8,8) & 0\tabularnewline
\hline 
Conv2D & (128,8,8) & 73856\tabularnewline
\hline 
Dropout & (128,8,8) & 0\tabularnewline
\hline 
Conv2D & (128,8,8) & 147584\tabularnewline
\hline 
MaxPooling2D & (128,4,4) & 0\tabularnewline
\hline 
Flatten & (2048) & 0\tabularnewline
\hline 
Dropout & (2048) & 0\tabularnewline
\hline 
Dense & (1024) & 2098176\tabularnewline
\hline 
Dropout & (1024) & 0\tabularnewline
\hline 
Dense & (512) & 524800\tabularnewline
\hline 
Dropout & (512) & 0\tabularnewline
\hline 
Dense & (10) & 5130\tabularnewline
\hline 
\end{tabular}
\par\end{centering}
}
\par\end{centering}
\begin{centering}
\subfloat[Network architecture for MNIST.]{\begin{centering}
\begin{tabular}{|c|c|c|}
\hline 
Layer type & Output Shape & Param \#\tabularnewline
\hline 
\hline 
Conv2D & (32,24,24) & 832\tabularnewline
\hline 
MaxPooling2D & (32,12,12) & 0\tabularnewline
\hline 
Dropout & (32,12,12) & 0\tabularnewline
\hline 
Flatten & (4608) & 0\tabularnewline
\hline 
Dense & (128) & 589952\tabularnewline
\hline 
Dense & (10) & 1290\tabularnewline
\hline 
\end{tabular}
\par\end{centering}
}
\par\end{centering}
\selectlanguage{british}%
\selectlanguage{british}%
\end{table}
\selectlanguage{british}%

\end{document}